# Natural Language Processing: The Acquisition of Semantic Relationships

**Mohamed Naamane**
**University Ibn Khaldoun, DZ**

**Abstract**

The study of semantic relationships has revealed a close connection between these relationships and the morphological characteristics of a language. Morphology, as a subfield of linguistics, investigates the internal structure and formation of words. By delving into the relationship between semantic relationships and language morphology, we can gain deeper insights into how the underlying structure of words contributes to the interpretation and comprehension of language. This paper explores the dynamic interplay between semantic relationships and the morphological aspects of different languages, by examining the intricate relationship between language morphology and semantic relationships, valuable insights can be gained regarding how the structure of words influences language comprehension.

## 1. Introduction

The fundamental premise of this algorithm rests upon the extraction of a value that effectively captures the intrinsic relationship between two given words. This value serves as a pivotal seed in the discovery of analogous relationships with other words. It is noteworthy that this approach to numerical representation is not a recent development; rather, it finds its historical roots in the practices of ancient Arab civilizations.

## 2. Abjad_numerals

The assignment of numerical values to individual English letters follows a prescribed methodology:

DB["a"]=1

DB['b']=2

DB['c']=3

DB['d']=4

DB["e"]=5

DB["f"]=6

DB["g"]=7

DB["h"]=8

DB["i"]=9

DB["j"]=10

DB["k"]=20

DB["l"]=30

DB["m"]=40

DB["n"]=50

DB["o"]=60

DB["p"]=70

DB["q"]=80

DB["r"]=90

DB["s"]=100

DB["t"]=200

DB["u"]=300

DB["v"]=400

DB["w"]=500

DB["x"]=600

DB["y"]=700

DB["z"]=800

## 3. Semantic Relationships

The extraction of a value represents the relationship between two given words, denoted as x and y, entails a process of converting these words into series of numbers based on the assigned numerical values of their constituent letters.

$$S(x) = \sum_{i=0}^{n-1} S(\sum_{j=0}^{n-1} S(xi))$$

As an illustration, let's consider the value of the letter "t" to be 200. In this case, the corresponding series of numbers for the word containing only the letter "t" would be represented as [2, 0, 0], where each number represents a single digit. To ensure equal series of numbers sizes between x and y, if the size of the series of numbers for x is smaller than the series of numbers for y, zeros are appended to the series of numbers for x until both series of numbers attain equal length.

The average product obtained by dividing the sum of the multiplication results between the corresponding elements within the two series of numbers by 10 serves as a quantitative measure that expresses the relationship between the two words. This value, often referred to as the seed, can be utilized to infer analogous relationships with other words. Specifically, it enables the establishment of a ratio that signifies the strength of the relationship between words x and z:

μ = ∑(|A⊙B|) / n

R(x,y) = μ/10

seed = R(x,y)

To explore additional words that share a similar relationship as the previously established relationship between x and y, we will compute a value that signifies the strength of the relationship between the seed relation and another word, denoted as z. This value serves as a metric to gauge the degree of association between x and z, allowing for the identification of words that exhibit analogous relationships to the initial pair:

R(x,z) = [|seed * R(x,z) * 100|]

This is an illustrative example of applying this algorithm to examine the relationship between the word pair ['brain', 'think'] within a small English corpus. The analysis shows that the word "imaging" exhibits the highest degree of correlation with a 237% relevance score after the word Brain:

```
# git clone -b master
https://github.com/mrmednmn/wre.git

from wre.lang.arabic import Arabic
from wre.lang.english import English
from wre.wrelation import WordRelation

lang = English()
word_relation = WordRelation(lang)
word_pair = ["brain","think"]
#get the relation between 2 words as a seed, the word pair should have a relation.
seed = word_relation.get_relation(word_pair[0], word_pair[1])

#input file to get similar relations from it
data = open("tkns.txt", "r", encoding=lang.encoding).read().split()
total_words = len(data)

with open("out.txt", "w", encoding=lang.encoding) as out_file:
    for index, word in enumerate(data):
        progress_percentage = (index + 1) / total_words * 100
        progress_message = f"Processing word {index + 1} of {total_words} ({progress_percentage:.2f}%)"
        print(progress_message, end='\r', flush=True)
        #ge the relation between the seed relation and the curent word in the input data
        relation = word_relation.get_relation(word_pair[0], word)
        #check if the curr input word has a relation between the seed word pair
        if word_relation.has_relation(seed, relation):
            #get the diffrence between the seed and the curent relation
            diff=word_relation.evaluate(seed, relation)
            response = str(word_pair) + " is related to " + word +" per = " + str(diff)+"%"
            out_file.write(response+"\n")
        clear_message = " " * len(progress_message)
        print(clear_message, end='\r')
```

This is an illustrative example of applying this algorithm to examine the relationship between the word pair ['brain', 'think'] within a small English corpus. The analysis shows that the word "imaging" exhibits the highest degree of correlation with a 237% relevance score after the word "Brain".

### 4.Results

['brain', 'think'] is related to university per = 33%
['brain', 'think'] is related to book per = 139%
['brain', 'think'] is related to wave per = 77%

['brain', 'think'] is related to ai per = 155%
['brain', 'think'] is related to technology per = 55%
['brain', 'think'] is related to lesson per = 7%
['brain', 'think'] is related to succuss per = 4%
['brain', 'think'] is related to tree per = 71%
['brain', 'think'] is related to head per = 122%
['brain', 'think'] is related to black per = 129%
['brain', 'think'] is related to food per = 172%
['brain', 'think'] is related to sleep per = 44%
['brain', 'think'] is related to learn per = 144%
['brain', 'think'] is related to study per = 4%
['brain', 'think'] is related to class per = 56%
['brain', 'think'] is related to sport per = 43%
['brain', 'think'] is related to happy per = 128%
['brain', 'think'] is related to prepare per = 112%
['brain', 'think'] is related to romantic per = 51%
['brain', 'think'] is related to people per = 72%
['brain', 'think'] is related to Cognition per = 98%
['brain', 'think'] is related to Neurology per = 9%
['brain', 'think'] is related to Intelligence per = 47%
['brain', 'think'] is related to Memory per = 46%
['brain', 'think'] is related to Cognitive per = 98%
['brain', 'think'] is related to functions per = 61%
['brain', 'think'] is related to Neuroplasticity per = 6%
['brain', 'think'] is related to Synapse per = 8%
['brain', 'think'] is related to Neural per = 15%
['brain', 'think'] is related to networks per = 9%
['brain', 'think'] is related to Neurotransmitters per = 5%
['brain', 'think'] is related to Cognitive per = 98%
['brain', 'think'] is related to development per = 51%
['brain', 'think'] is related to Brain per = 360%
['brain', 'think'] is related to structure per = 2%
['brain', 'think'] is related to Brain per = 360%
['brain', 'think'] is related to activity per = 61%
['brain', 'think'] is related to Neural per = 15%
['brain', 'think'] is related to pathways per = 12%
['brain', 'think'] is related to Cognitive per = 98%
['brain', 'think'] is related to processes per = 50%
['brain', 'think'] is related to Mental per = 27%
['brain', 'think'] is related to processes per = 50%
['brain', 'think'] is related to Brain per = 360%
['brain', 'think'] is related to health per = 108%
['brain', 'think'] is related to Brain per = 360%
['brain', 'think'] is related to functions per = 61%
['brain', 'think'] is related to Neurological per = 9%
['brain', 'think'] is related to disorders per = 98%
['brain', 'think'] is related to Brain per = 360%
['brain', 'think'] is related to imaging per = 237%
['brain', 'think'] is related to Neurodegeneration per = 6%

### 5. Conclusion

In conclusion, the investigation into the relationship between language morphology and semantic associations has shed light on the potential for extracting valuable insights from the numerical representation of words. By employing a methodology that involves assigning numerical values to letters and subsequently generating series of numbers we have established a means to quantify the strength of relationships between words. This approach, exemplified by the computation of average products and ratios, enables the inference of analogous relationships and the discovery of words that share similar semantic connections.

As we continue to explore the complex relationship between language morphology and semantic associations, further research endeavors should focus on refining the computational techniques employed, expanding the linguistic datasets utilized, and investigating the generalizability of these findings across diverse languages and linguistic contexts. By doing so, we can continue to advance our understanding of language systems and unlock new avenues for linguistic analysis and application.

### 6. References

[1] Martijn Bartelds, Nay San, Bradley McDonnell, Dan Jurafsky and Martijn Wieling. 2023. Making More of Little Data: Improving Low-Resource Automatic Speech Recognition Using Data Augmentation. To appear in ACL 2023.

[2] Mirac Suzgun, Luke Melas-Kyriazi, Dan Jurafsky. 2023. Follow the Wisdom of the Crowd: Effective Text Generation via Minimum Bayes Risk Decoding. Draft to appear in Findings of ACL 2023.

[3] https://en.wikipedia.org/wiki/Abjad_numerals

[4] Eugenia H. Rho, Maggie Harrington, Yuyang Zhong, Reid Pryzant, Nicholas P. Camp, Dan Jurafsky, and Jennifer L. Eberhardt. 2023. Escalated police stops of Black men are linguistically and psychologically distinct in their earliest moments. PNAS 120 (23).